\pdfoutput=1
\documentclass[11pt]{article}

\usepackage{EMNLP2023}

\usepackage{times}
\usepackage{latexsym}

\usepackage[T1]{fontenc}
\usepackage[utf8]{inputenc}
\usepackage{inconsolata}

\usepackage{booktabs}
\usepackage{graphicx}
\usepackage{amssymb}
\usepackage{amsmath}
\usepackage{float}
\usepackage{enumitem}
\usepackage{multirow}
\usepackage{xcolor}

\usepackage[most]{tcolorbox}
\newtcolorbox{casebox}[1][]{%
    enhanced,
    colback=white,
    colframe=gray!50,
    coltitle=black,
    fonttitle=\bfseries\small\sffamily,
    boxrule=0.8pt,
    drop shadow=black!15,
    sharp corners,
    rounded corners=southeast,
    arc is angular,
    arc=3mm,
    title={#1},
    attach boxed title to top left={yshift=-2mm, xshift=2mm},
    boxed title style={colback=white, frame hidden},
    top=6pt,
    bottom=6pt,
    left=6pt,
    right=6pt
}

\usepackage{tikz}
\usepackage{pgfplots}
\pgfplotsset{compat=1.18}
\usetikzlibrary{
    shapes.geometric,
    arrows.meta,
    positioning,
    shadows.blur,
    backgrounds,
    fit,
    calc,
    decorations.pathreplacing,
    shapes.symbols,
    patterns
}

\pgfdeclarelayer{background}
\pgfdeclarelayer{foreground}
\pgfsetlayers{background,main,foreground}

\definecolor{ssgmblue}{RGB}{0,85,160}
\definecolor{ssgmred}{RGB}{180,30,30}
\definecolor{ssgmgreen}{RGB}{30,130,60}
\definecolor{bgblue}{RGB}{240,245,255}
\definecolor{bgred}{RGB}{255,245,245}
\definecolor{bggray}{RGB}{250,250,250}

\title{Governing Evolving Memory in LLM Agents: Risks, Mechanisms, and the\\ Stability and Safety Governed Memory (SSGM) Framework}
\author{
  Chingkwun Lam, Jiaxin Li, Lingfei Zhang, Kuo Zhao\thanks{\ \ Corresponding author.}  \\
  College of Intelligent Science and Engineering, Jinan University \\
  \texttt{zhaokuo@jnu.edu.cn}
}
\date{}

\begin{document}
\maketitle

\begin{abstract}
Long-term memory has emerged as a foundational component of autonomous Large Language Model (LLM) agents, enabling continuous adaptation, lifelong multimodal learning, and sophisticated reasoning. However, as memory systems transition from static retrieval databases to dynamic, agentic mechanisms, critical concerns regarding memory governance, semantic drift, and privacy vulnerabilities have surfaced. While recent surveys have focused extensively on memory retrieval efficiency, they largely overlook the emergent risks of memory corruption in highly dynamic environments. To address these emerging challenges, we propose the Stability and Safety-Governed Memory (SSGM) framework, a conceptual governance architecture. SSGM decouples memory evolution from execution by enforcing consistency verification, temporal decay modeling, and dynamic access control prior to any memory consolidation. Through formal analysis and architectural decomposition, we show how SSGM can mitigate topology-induced knowledge leakage where sensitive contexts are solidified into long-term storage, and help prevent semantic drift where knowledge degrades through iterative summarization. Ultimately, this work provides a comprehensive taxonomy of memory corruption risks and establishes a robust governance paradigm for deploying safe, persistent, and reliable agentic memory systems.
\end{abstract}

\section{Introduction}
Large Language Model (LLM) agents have demonstrated impressive reasoning and interaction capabilities across diverse domains \citep{MatarazzoTorlone2025}; yet, in the absence of specialized mechanisms, they remain fundamentally \textit{stateless}. Standard LLMs rely on a fixed-length context window that prevents indefinite information retention \citep{ZhongEtAl2023MemoryBank, YousufEtAl2024, OpenAI2023GPT4Card}. While early solutions employed Retrieval-Augmented Generation (RAG) to provide a static knowledge base \citep{LewisEtAl2020RAG, GaoEtAl2023RAG}, modern autonomous agents demand a more dynamic capability: the ability to learn from experience, update their world models, and refine their strategies over time \citep{XiEtAl2023Rise, WangEtAl2023Voyager}.

This necessity has driven a paradigm shift from static memory storage to \textit{adaptive, self-refining memory systems}. Recent architectures treat memory operations not as passive retrieval tasks but as active decision-making processes. For instance, \textbf{Memory-R1} employs reinforcement learning to train specialized sub-agents that autonomously decide when to add, update, or delete memory units based on task feedback \citep{YanEtAl2025MemoryR1}. Similarly, frameworks like \textbf{Mem0} and \textbf{AtomMem} introduce dynamic consolidation mechanisms that continuously optimize the storage structure via atomic operations \citep{ChhikaraEtAl2025Mem0, HuoEtAl2026AtomMem}. In these systems, memory is no longer an immutable log but a mutable asset that evolves alongside the agent \citep{ZhangEtAl2025cMemAct, ZhangEtAl2026EvolutionSurvey}.

\begin{figure}[t]
\centering
\includegraphics[width=\linewidth]{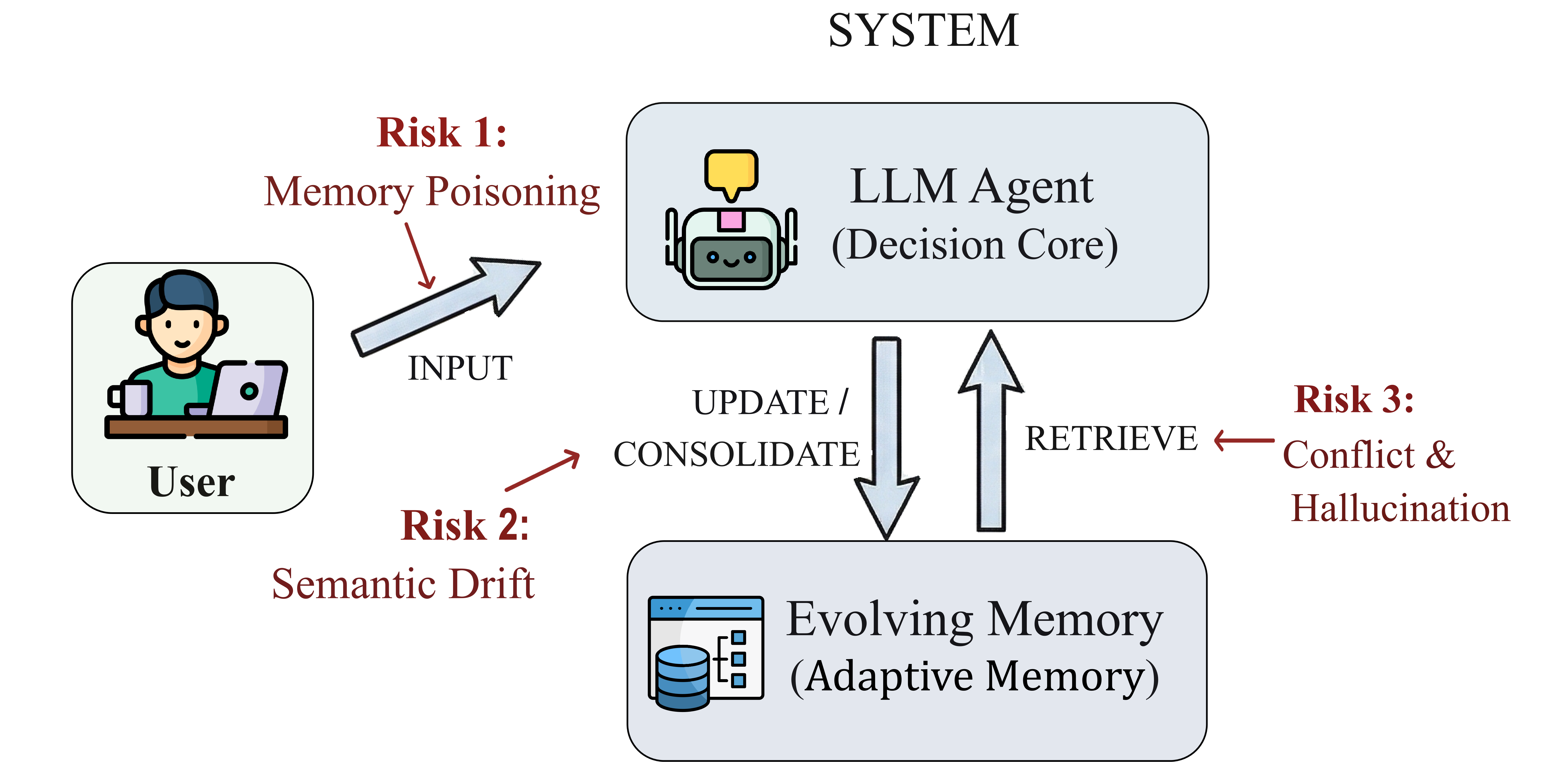}
\caption{\textbf{The Lifecycle of Memory Evolution and Emergent Risks.} Unlike static RAG, evolving memory systems introduce a feedback loop where errors can accumulate. We identify three critical failure points: (1) \textbf{Memory Poisoning} during input ingestion, (2) \textbf{Semantic Drift} during consolidation updates, and (3) \textbf{Conflict/Hallucination} during retrieval. SSGM aims to govern these interfaces.}
\label{fig:evolution_risks}
\end{figure}

However, granting agents the autonomy to rewrite their own memory introduces the \textbf{stability-plasticity dilemma} into artificial systems. Without robust governance, the continuous refinement of memory creates significant risks. An agent may gradually distort facts through repeated summarization (\textit{semantic drift}), reinforce suboptimal workflows (\textit{procedural drift}) \citep{HanEtAl2025LEGOMem, RathEtAl2026AgentDrift}, or inadvertently internalize hallucinations and malicious injections as valid knowledge \citep{XiangEtAl2025PrivacyRisks, GreshakeEtAl2023Injection}. Unlike static RAG systems, where errors are isolated to a single retrieval step, errors in evolving memory systems are cumulative and persistent. As delineated in Figure~\ref{fig:evolution_risks}, this creates a compounding failure loop across three critical interfaces: input ingestion (poisoning), memory consolidation (drift), and memory retrieval (hallucination). Figure~\ref{fig:evolution_risks} presents this coarse-grained lifecycle risk loop, which is later refined into the four-dimensional failure taxonomy in Table~\ref{tab:failures_expanded}. As noted in recent surveys \citep{HuEtAl2025AgeOfAgents, ZhangEtAl2026EvolutionSurvey}, while the mechanisms of memory update are well-studied, the protocols for ensuring their long-term correctness and safety remain underexplored.

To address this gap, this paper proposes the \textbf{Stability- and Safety-Governed Memory (SSGM)} framework. We argue that for LLM agents to be reliable in high-stakes environments, memory evolution must be decoupled from memory governance. Our contributions are fourfold:
\begin{enumerate}
    \item \textbf{Taxonomy of Evolution:} We categorize memory evolution along three dimensions: content abstraction, structural reorganization (e.g., from lists to Zettelkasten-style graphs \citep{XuEtAl2025AMem, JiangEtAl2026MAGMA}), and policy optimization.
    \item \textbf{Failure Analysis:} We identify and formalize key failure modes in adaptive memory, specifically distinguishing between intrinsic drift (e.g., knowledge conflict) and extrinsic threats (e.g., memory poisoning).
    \item \textbf{The SSGM Framework:} We synthesize design principles for a governed memory architecture that integrates consistency verification and ground-truth anchoring to mitigate the risks of uncontrolled evolution.
    \item \textbf{Fundamental Trade-offs:} We establish a formal discussion on the three fundamental trade-offs inherent in agentic memory---the latency-safety trade-off, the stability-plasticity conflict, and the scalability of graph structures---setting the stage for future research.
\end{enumerate}

\section{From Static to Adaptive Memory Systems}
Initial iterations of LLM-based agent systems either lacked long-term memory entirely or relied on simplistic approaches to maintain context. For example, some agents maintained a sequential log of recent dialogue and instructions, truncating or discarding older content upon the saturation of the context window \citep{PressEtAl2022ALiBi}; others utilized vector-similarity search within a database of past interactions to retrieve relevant segments on demand. These strategies constitute a paradigm of \textit{static} memory: the agent does not alter its memory usage adaptively based on interaction outcomes, and the rules for storage and retrieval are predetermined by developers. A classic example is the \textbf{Generative Agents} framework \citep{ParkEtAl2023GenerativeAgents}, where simulated characters stored factual observations and reflections, retrieving them via a fixed relevance-and-recency scoring heuristic. While effective to a degree, this design required extensive manual tuning of storage and summarization criteria, and the system could not learn from memory management errors. Similarly, \textbf{MemoryBank} \citep{ZhongEtAl2023MemoryBank} and related systems summarized or discarded observations using fixed schedules or manually crafted triggers, illustrating the resource-intensive and suboptimal nature of heuristic memory management.

The limitations of static approaches motivated the development of \textit{adaptive} memory mechanisms. A crucial insight involved framing memory management as a decision problem amenable to learning or operating system-like management. \textbf{MemGPT} \citep{PackerEtAl2023MemGPT} pioneered the concept of treating context as a limited resource, employing an OS-inspired paging mechanism to transfer information between ``main context'' (RAM) and ``external context'' (disk) based on immediate task requirements. This represented a fundamental shift from simple sliding windows. \textbf{The Tensor Brain} \citep{TrespEtAl2023TensorBrain} constituted another early effort, introducing a neural architecture for differentiable episodic, semantic, and working memory. More recently, Memory-R1 explicitly formulates memory operations as actions to be optimized by a learned policy \citep{YanEtAl2025MemoryR1}.

Other systems incorporate adaptivity via alternative mechanisms: \textbf{MemAct} \citep{ZhangEtAl2025cMemAct} enables the agent to embed memory-related actions into its chain-of-thought prompting \citep{WeiEtAl2022CoT, YaoEtAl2023ReAct}. If the agent detects uncertainty regarding a detail, it can explicitly issue a ``lookup memory'' or ``store memory'' action. Over multiple reasoning sessions, the agent refines the timing of these actions via feedback. Similarly, \textbf{Nemori} \citep{NanEtAl2025Nemori} draws upon cognitive science to introduce a self-organizing memory policy: the agent internalizes rules for maintaining consistency and organization (such as periodically reconciling new information with old) without external supervision, thereby gradually improving memory coherence.

\begin{table*}[t]
\centering
\small
\resizebox{\textwidth}{!}{%
\begin{tabular}{@{}llllll@{}}
\toprule
\textbf{System} & \textbf{Memory Structure} & \textbf{Evolution Policy} & \textbf{Refinement \& Stability} & \textbf{Safety \& Access} & \textbf{Target Domain} \\
\midrule
\multicolumn{6}{c}{\textit{Adaptive \& Learning-Based Systems}} \\
\midrule
\textbf{Memory-R1} \citep{YanEtAl2025MemoryR1} & Flat Vector DB & \textbf{RL (PPO)} & Feedback-driven Updates & -- & Open-Domain QA \\
\textbf{MemAgent} \citep{YuEtAl2025MemAgent} & Semantic Slots & \textbf{RL (DAPO)} & Selective Overwriting & -- & Long-Context QA \\
\textbf{AtomMem} \citep{HuoEtAl2026AtomMem} & Vector + Buffer & Atomic Ops & Needle-in-Haystack Training & -- & Multi-Turn Reasoning \\
\textbf{LEGOMem} \citep{HanEtAl2025LEGOMem} & \textbf{Procedural Graph} & Failure-driven & Rule Verification & -- & Workflow Automation \\
\textbf{AgentSM} \citep{BiswalEtAl2026AgentSM} & SQL-based Knowledge & Agentic Tuning & Schema Evolution & Strict Schema Constraints & Text-to-SQL (Enterprise) \\
\textbf{AWM} \citep{ZhengEtAl2024AWM} & Workflow Rules & \textbf{Skill Induction} & Trace Compression & -- & Procedural Execution \\
\textbf{DarwinMem} \citep{MiEtAl2026DarwinMem} & Evolutionary Pool & \textbf{Training-Free} & Survival of the Fittest & -- & GUI Agent Evolution \\
\textbf{Astraea} \citep{NiEtAl2025Astraea} & KV Cache & State-Aware Sched. & JCT Lifecycle Optim. & -- & Agentic Workflows \\
\midrule
\multicolumn{6}{c}{\textit{Graph-Based \& Cognitive Architectures}} \\
\midrule
\textbf{HiMem} \citep{ZhangEtAl2026HiMem} & \textbf{Hierarchical} & LLM-Reasoning & \textbf{Reconsolidation} & Precision-Recall Balance & Long-Horizon Agents \\
\textbf{A-MEM} \citep{XuEtAl2025AMem} & \textbf{Zettelkasten Graph} & Self-Organizing & Dynamic Linking & -- & General Agents \\
\textbf{MemoRAG} \citep{QianEtAl2024MemoRAG} & \textbf{Global Memory Graph} & Insight Extraction & Memory-Enhanced Retrieval & -- & Long Context QA \\
\textbf{LiCoMemory} \citep{HuangEtAl2025LiCoMemory} & Lightweight KG & Consistency Checks & \textbf{Semantic Decay} & Fact Verification & Multi-Turn Reasoning \\
\textbf{HippoRAG} \citep{Gutierrez2024HippoRAG} & Neurobiological Graph & Spreading Act. & Pathway Consolidation & -- & Open-Domain QA \\
\textbf{Forgetful/Faithful} \citep{Alqithami2025Forgetful} & Typed Node Schema & Budgeted Forgetting & Coherence Preserved & \textbf{$(\epsilon,\delta)$-Privacy} & Privacy-Aware Agents \\
\textbf{E-mem} \citep{WangEtAl2026EMem} & Episodic Context & Reconstruction & Context Reconstruction & Traceability & Long-Horizon Tasks \\
\textbf{Zep} \citep{RasmussenEtAl2025Zep} & Temporal KG & Incremental Extraction & Entity Resolution & -- & Conversational AI \\
\midrule
\multicolumn{6}{c}{\textit{Multimodal \& Collaborative Systems}} \\
\midrule
\textbf{TeleMem} \citep{ChenEtAl2025TeleMem} & Object-Centric Graph & User-driven Trigger & Physical Scene Updates & Privacy Filtering & \textbf{Embodied AI / IoT} \\
\textbf{VideoARM} \citep{YinEtAl2025VideoARM} & Hierarchical Multi. & Adaptive Loop & Multi-level Clue Updates & -- & Long-Form Video \\
\textbf{WorldMM} \citep{YeoEtAl2025WorldMM} & Spatiotemporal Mem & Dynamic Retrieval & Video State Tracking & -- & Long Video Reasoning \\
\textbf{MemVerse} \citep{LiuEtAl2025MemVerse} & Multimodal Vector & Continual Learning & Cross-Modal Alignment & -- & Lifelong Learning \\
\textbf{Collab. Mem} \citep{RezazadehEtAl2025CollaborativeMemory} & Distributed Vector & Consensus Voting & Cross-User Deduplication & \textbf{Dynamic ACLs} & Multi-User Collab \\
\textbf{Topology Matters} \citep{LiuEtAl2025Topology} & Multi-Agent Network & MAMA Framework & Leakage Plateau Anal. & \textbf{Centrality Const.} & Multi-Agent Privacy \\
\textbf{MIRIX} \citep{WangChen2025MIRIX} & Shared Multi-Agent Space & Collaborative Reflection & Procedural Alignment & Agent Role Boundaries & Multi-Agent Systems \\
\bottomrule
\end{tabular}%
}
\caption{\textbf{Taxonomy of Evolving Memory Systems.} We categorize recent systems by their storage structure, the policy governing memory evolution, mechanisms for maintaining stability, and safety provisions. Newly integrated approaches like \textbf{HippoRAG}, \textbf{DarwinMem}, and \textbf{VideoARM} emphasize the shift towards neurobiology, adaptive policies, and multimodal environments.}
\label{tab:systems_comparison}
\end{table*}

\section{What Evolves: Content, Structure, and Policy}
Memory in LLM agents evolves along multiple dimensions. We categorize the major facets as: (1) the \textit{content} of memory, comprising stored information and its level of detail; (2) the \textit{structure} of memory, referring to the organization and representation of knowledge; and (3) the \textit{policy} for memory management, namely the decision-making process that determines how memory is utilized and updated. While these facets are interrelated, examining them separately clarifies how modern systems transcend the limitations of static memory.

\subsection{Evolving Memory Content (What is Stored)}
The most direct form of memory evolution occurs in the content itself. Unlike a static transcript, adaptive agents continually modify memory units. This process involves adding new facts, updating existing units, summarizing detailed events into high-level conclusions, or deleting information deemed irrelevant or obsolete. By altering the \textit{content} of stored information, the knowledge base becomes a dynamic resource that changes over time.

A common pattern is \textbf{summarization and abstraction}. An agent may replace a detailed episode with a brief synopsis once confidence is established that fine-grained details are no longer requisite. This was observed in early experiments with Generative Agents \citep{ParkEtAl2023GenerativeAgents}. More recently, systems have expanded content evolution to include \textbf{multimodal and procedural knowledge}. For instance, \textbf{VideoARM} \citep{YinEtAl2025VideoARM} dynamically constructs hierarchical memory to reason over long-form videos without exhaustive preprocessing, while \textbf{WorldMM} \citep{YeoEtAl2025WorldMM} continuously tracks spatiotemporal states. Similarly, \textbf{TeleMem} \citep{ChenEtAl2025TeleMem} maintains evolving object-centric graphs that track the changing states of physical environments, while \textbf{LEGOMem} \citep{HanEtAl2025LEGOMem} synthesizes successful execution traces into procedural rules. To further structure this, frameworks like \textbf{Agent Workflow Memory (AWM)} \citep{ZhengEtAl2024AWM} demonstrate how episodic traces can be systematically compressed into reusable workflow rules, enhancing procedural memory while mitigating the noise of raw logs. This shift from purely textual facts to procedural content increases the complexity of maintaining consistency, as errors in procedural memory can lead to cascading failures in task execution.

\subsection{Evolving Memory Structure (How Knowledge is Organized)}
The structuring of memory dramatically influences retrieval effectiveness and the reasoning capabilities of the agent \citep{ZhangEtAl2026EvolutionSurvey}. Grouping related facts or linking causes and effects facilitates inferences that are difficult to derive from unstructured data. Many agent memory systems draw inspiration from databases or cognitive metaphors such as knowledge graphs and semantic networks.

Recent advancements have introduced self-organizing structures. \textbf{A-MEM} \citep{XuEtAl2025AMem} adopts the \textit{Zettelkasten} method, treating memories as atomic notes that dynamically evolve by establishing links with related concepts, effectively creating a ``living'' graph of knowledge. Drawing directly from neurobiology, \textbf{HippoRAG} \citep{Gutierrez2024HippoRAG} utilizes a spreading activation mechanism over knowledge graphs to mimic human associative recall, drastically improving multi-hop reasoning performance. To bridge the gap between static retrieval and cognitive synthesis, \textbf{MemoRAG} \citep{QianEtAl2024MemoRAG} builds a global memory graph over long contexts, enabling agents to extract structural insights before generating answers. Similarly, \textbf{HiMem} \citep{ZhangEtAl2026HiMem} implements a hierarchical structure where episodic details are continuously distilled into semantic knowledge through a process of \textit{reconsolidation}. \textbf{ChatDB} \citep{ChenEtAl2024ChatDB} treats memory as a symbolic SQL database, allowing the agent to execute complex queries rather than relying solely on semantic search; the structure evolves as the agent defines new tables or schemas.

\subsection{Evolving Memory Policy (How Memory is Managed)}
\textbf{Table~\ref{tab:systems_comparison} provides a comprehensive taxonomy} of these evolving systems, categorizing them by their structural approach (from vector databases to hierarchical graphs) and their evolution policies. Notably, as shown in the table, newer frameworks like \textbf{AtomMem} \citep{HuoEtAl2026AtomMem} and \textbf{MemAgent} \citep{YuEtAl2025MemAgent} employ Reinforcement Learning (RL) to learn optimal memory policies. By framing memory operations (add, update, delete, retrieve) as atomic actions, these agents optimize their memory management strategies to maximize long-term task rewards. Conversely, \textbf{DarwinMem} \citep{MiEtAl2026DarwinMem} proves that memory policies can also be optimized without parameter updates, employing a training-free ``survival of the fittest'' evolutionary pool to adapt GUI agents across tasks.

\section{How Memory Evolves: Reflection, Reinforcement, and Consolidation}
Having identified \textit{what} changes in evolving memory, we now examine \textit{how} these changes are effected.

\subsection{Reflection and Self-Supervised Memory Updates}
\textbf{Reflection} involves an agent examining its own actions or outcomes and generating analyses or lessons, which then inform future decisions \citep{SumersEtAl2024Cognitive}. In Reflexion \citep{ShinnEtAl2023Reflexion}, the agent prompts itself: ``If the solution was wrong, explain the error.'' The generated reflection is stored as a memory. This reflective loop has become a cornerstone of agent improvement \citep{MadaanEtAl2023SelfRefine, WangEtAl2023Voyager}. \textbf{MemR$^3$} \citep{DuEtAl2025MemR3} extends this by employing reflective reasoning during the retrieval process itself, refining queries iteratively to bridge the gap between user intent and stored knowledge.

\subsection{Reinforcement Learning and Outcome-Driven Memory Optimization}
While reflection relies on self-supervision, \textbf{Reinforcement Learning (RL)} offers a direct method to optimize memory behavior toward explicit objectives. Formally, the memory management problem can be modeled as a Partially Observable Markov Decision Process (POMDP). At time $t$, the agent receives an observation $o_t$ (e.g., a user query), maintains an internal memory state $M_t$, and occupies a latent decision state $s_t$ that summarizes the observation, the current memory, and the task context. The policy $\pi_\theta(a_t \mid o_t, M_t)$ selects an action $a_t \in \mathcal{A}$, where the global action space contains both task-level actions and memory-management actions. We explicitly define the memory-action subset as follows:
\begin{equation}
\begin{split}
\mathcal{A}_{\text{mem}} = \{ & \textsc{Add}(c), \textsc{Update}(idx, v), \\
& \textsc{Delete}(idx), \textsc{Retrieve}(q) \}
\end{split}
\end{equation}
where $c$ denotes the content to be stored, $idx$ denotes the index or key of an existing memory item, $v$ denotes the replacement or updated value written to that indexed item, and $q$ denotes the retrieval query used to access previously stored memory. Intuitively, Eq.~(1) defines the discrete action subspace through which the agent treats memory management as an explicit decision process rather than as a fixed backend routine.

The agent optimizes parameters $\theta$ to maximize the expected discounted return:
\begin{equation}
J(\theta) = \mathbb{E}_{\tau \sim \pi_\theta} \left[ \sum_{t=0}^{T} \gamma^t r(s_t, a_t) \right]
\end{equation}
where $\tau$ denotes a trajectory induced by the policy $\pi_\theta$, $T$ denotes the horizon length of the interaction episode, $\gamma \in [0,1)$ is the discount factor that balances immediate and long-term reward, $s_t$ denotes the latent decision state at time $t$, and $r(s_t,a_t)$ denotes the reward assigned to taking action $a_t$ in state $s_t$. Intuitively, Eq.~(2) formalizes the objective of learning a memory policy that optimizes long-term utility: a memory write that incurs slight short-term cost may still be preferred if it substantially improves future reasoning or retrieval. Memory-R1 \citep{YanEtAl2025MemoryR1} utilizes this formulation via Proximal Policy Optimization (PPO). \textbf{MemAgent} \citep{YuEtAl2025MemAgent} further refines this by using Group Relative Policy Optimization (GRPO) to handle the sparse rewards associated with long-term memory benefits.

\subsection{Memory Consolidation and Forgetting}
The third key mechanism is \textbf{consolidation and forgetting}, analogous to biological processes \citep{TrespEtAl2023TensorBrain}. Without management, the memory size $|\mathcal{M}|$ grows linearly, causing retrieval latency. \citet{Alqithami2025Forgetful} formalizes this through ``forgetting-by-design'', demonstrating via the FiFA benchmark that bounded, budget-aware forgetting policies (e.g., Priority Decay) not only reduce computational cost but actively preserve narrative coherence and privacy without sacrificing functionality.

\paragraph{Temporal Decay and Freshness Governance}
To mitigate the risk of \textit{temporal obsolescence}, where stale facts contradict recent updates, governance mechanisms must incorporate time-aware forgetting. Inspired by cognitive decay theories, \citet{HuangEtAl2025LiCoMemory} propose a Weibull-based decay function to model the relevance $w(\Delta \tau)$ of a memory unit over time:
\begin{equation}
    w(\Delta \tau) = \exp \left( - \left( \frac{\Delta \tau}{\eta} \right)^\kappa \right)
\end{equation}
where $\Delta \tau$ denotes the elapsed time (e.g., measured in conversational turns or absolute hours) since the last successful retrieval of the memory unit, $\eta$ is the scale parameter controlling the characteristic time scale of decay, and $\kappa$ is the shape parameter controlling the curvature of the forgetting profile. Intuitively, Eq.~(3) assigns lower relevance scores to stale memory entries, thereby allowing the system to downweight or discard items that are less likely to remain valid. The Weibull form is more expressive than a simple exponential decay because the shape parameter can represent faster early decay or more delayed forgetting under different domains. In the SSGM framework, memory items are pruned or archived whenever their temporal relevance falls below a unified freshness threshold $\theta_{\text{fresh}}$, thereby reducing the attack surface for stale reasoning and memory-induced hallucination.

Furthermore, \textbf{E-mem} \citep{WangEtAl2026EMem} shifts from simple retrieval to \textit{Episodic Context Reconstruction}, where the agent actively reconstructs the context from fragmented traces rather than retrieving pre-stored blocks, ensuring higher fidelity to the original experience.

\section{Why Memory Fails: Drift, Efficiency, Validity, and Safety}
Evolving memory empowers LLM agents but also introduces a spectrum of potential failure modes. As illustrated in Table~\ref{tab:failures_expanded}, we categorize these failures into four distinct dimensions: Stability, Validity, Efficiency, and Safety.

\begin{table*}[t]
\centering
\small
\resizebox{\textwidth}{!}{%
\begin{tabular}{@{}lllll@{}}
\toprule
\textbf{Category} & \textbf{Failure Mode} & \textbf{Mechanism \& Manifestation} & \textbf{Representative Contexts} & \textbf{SSGM Mitigation Strategy} \\
\midrule
\multirow{3}{*}{\textbf{Stability}}
 & \textbf{Semantic Drift} & Iterative summarization causes gradual nuance loss & Recursive Summarization \citep{ParkEtAl2023GenerativeAgents} & Ground Truth Anchoring ($\mathcal{R}$) \\
 & \textbf{Procedural Drift} & Reinforcement of suboptimal/outdated workflows & Workflow Automation \citep{HanEtAl2025LEGOMem,FangEtAl2025Memp} & Rule Verification \\
 & \textbf{Goal/Role Drift} & Alignment shift due to accumulated interaction bias & Long-term Role-play \citep{YuenEtAl2025IntrinsicMemory} & Role Partitioning \\
\midrule
\multirow{2}{*}{\textbf{Validity}}
 & \textbf{Memory Hallucination} & Retrieval of non-existent or fabricated facts & Open-domain QA \citep{DuEtAl2025MemR3} & Consistency Verifier (TMS) \\
 & \textbf{Temporal Obsolescence} & Conflict between stale memories and new states & User Personalization \citep{RasmussenEtAl2025Zep} & \textbf{Weibull Decay Function} \\
\midrule
\multirow{2}{*}{\textbf{Efficiency}}
 & \textbf{Retrieval Latency} & Search time scales linearly/quadratically with history & Real-time Interaction \citep{ChhikaraEtAl2025Mem0} & Hierarchical Indexing \\
 & \textbf{Index Bloat} & Accumulation of redundant/noisy episodic logs & Infinite Context \citep{SarinEtAl2025Memoria} & Active Forgetting / Pruning \\
\midrule
\multirow{2}{*}{\textbf{Safety}}
 & \textbf{Memory Poisoning} & Injection of malicious instructions into storage & User Instruction Tuning \citep{GreshakeEtAl2023Injection} & Write Filtering (Firewall) \\
 & \textbf{Privacy Leakage} & Unauthorized cross-session/cross-user retrieval & Multi-tenant Agents \citep{RezazadehEtAl2025CollaborativeMemory} & \textbf{Provenance \& ACLs} \\
\bottomrule
\end{tabular}%
}
\caption{\textbf{Comprehensive Taxonomy of Evolving Memory Failures.} We expand the taxonomy to include \textbf{Efficiency} failures (Latency, Bloat), which are critical in long-horizon agents. \textbf{Validity} failures now explicitly distinguish between hallucination and temporal obsolescence (fact conflict over time), citing recent evidence from \citet{RasmussenEtAl2025Zep}.}
\label{tab:failures_expanded}
\end{table*}

\subsection{Semantic Drift and Gradual Deviation}
\begin{figure}[t]
\centering
\includegraphics[width=6cm]{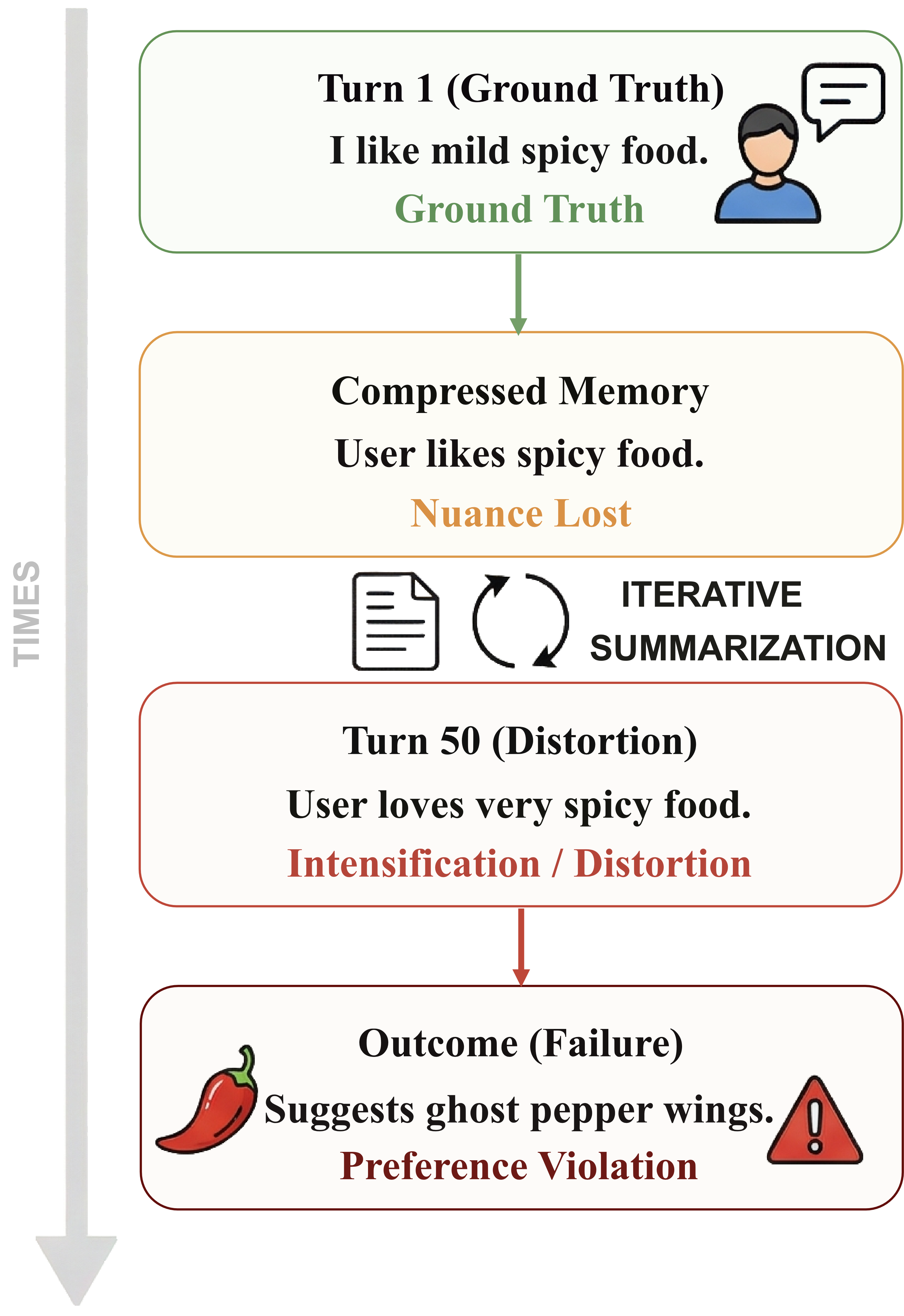}
\caption{\textbf{An Illustrative Case of Semantic Drift.} Iterative summarization gradually distorts a specific user preference through lossy compression and semantic intensification. For example, an originally mild preference (e.g., ``I like mild spicy food'') may be progressively rewritten as ``likes spicy food'' and later ``loves very spicy food,'' ultimately causing a preference violation (e.g., suggesting ghost pepper wings).}
\label{fig:drift_case}
\end{figure}

\textbf{Memory drift} refers to the phenomenon where the knowledge stored by an agent gradually deviates from ground truth (conceptually visualized in Figure~\ref{fig:drift_case}, which specifically illustrates preference-intensity drift under repeated lossy summarization). Drift is not monolithic; it manifests in several forms.

\textbf{Semantic drift} is primarily driven by \textbf{lossy compression algorithms}, such as iterative summarization. As information is re-encoded multiple times (e.g., from raw text to summary to higher-level reflection), nuanced details are stripped away, leading to a gradual distortion of the ground truth \citep{ParkEtAl2023GenerativeAgents}. Figure~\ref{fig:drift_case} provides an intuitive example of this process: a mild user preference can be progressively intensified through repeated rewriting, eventually causing a concrete preference violation. As contrasted in Figure~\ref{fig:drift_logic}, while a naive pipeline exacerbates this nuance loss by repeatedly overwriting the active memory, an anchored approach periodically reconciles the mutable state with an immutable episodic ledger (formally denoted as $\mathcal{K}_{\text{ledger}}$ later in Section 6.2) to preserve original fidelity. Figure~\ref{fig:drift_logic} abstracts the causal mechanism behind this process, whereas Figure~\ref{fig:drift_case} visualizes its concrete symptom at the level of a user preference.

\textbf{Procedural drift} refers to the degradation of the skills or strategies of an agent as it reinforces suboptimal execution paths \citep{HanEtAl2025LEGOMem}. For instance, an agent might ``learn'' a convoluted workaround for a simple API call and rigidify this into its procedural memory, blocking future optimization.

Finally, \textbf{Goal drift} represents a subtle misalignment where the behavior of the agent shifts away from original instructions due to accumulated biases in its memory \citep{RathEtAl2026AgentDrift}.

We can formalize semantic drift $\delta$ at time $T$ as the divergence between the embedding representation $E(M_T)$ of the current memory and the embedding of the ground-truth reference ledger $\mathcal{K}_{\text{true}}$:
\begin{equation}
\delta(M_T, \mathcal{K}_{\text{true}}) = 1 - \mathrm{sim}\left( E(M_T), E(\mathcal{K}_{\text{true}}) \right)
\end{equation}
where $M_T$ denotes the memory state after $T$ update steps, $\mathcal{K}_{\text{true}}$ denotes the idealized ground-truth semantic target, $E(\cdot) \in \mathbb{R}^d$ denotes a fixed semantic embedding model projecting the memory into a $d$-dimensional continuous space, and $\mathrm{sim}(\cdot,\cdot)$ denotes cosine similarity in the embedding space. Intuitively, Eq.~(4) uses embedding-space divergence as a tractable proxy for semantic corruption: the larger the distance between the current memory representation and the reference ledger, the larger the estimated drift. In practice, $\mathcal{K}_{\text{true}}$ may be approximated by an immutable ledger of raw observations, which preserves details that may be lost in iterative summarization.

\subsection{Validity Failures: Hallucination and Obsolescence}
Closely related to drift is the problem of \textbf{Validity}. This manifests primarily as \textbf{Memory Hallucination}, where the agent stores hallucinated content as truth. Distinct from hallucination is temporal obsolescence, a failure mode highlighted by \citet{RasmussenEtAl2025Zep} in temporal knowledge graphs. Here, stored information is factually correct but outdated. Without mechanisms to resolve conflicting timestamps (as addressed by LiCoMemory's decay function), agents may retrieve and act upon stale data.

\subsection{Systemic Failures: Efficiency and Safety}
Beyond correctness, evolving memory systems face Efficiency and Safety challenges. As memory logs expand, \textbf{Retrieval Latency} can scale linearly or quadratically \citep{ChhikaraEtAl2025Mem0}. \textbf{Astraea} \citep{NiEtAl2025Astraea} reveals that isolated segment retrieval causes severe Head-of-Line blocking, advocating for state-aware scheduling that optimizes the global request lifecycle during agentic memory operations.

On the safety front, \textbf{Memory Poisoning} remains a critical risk \citep{GreshakeEtAl2023Injection}. Additionally, in multi-user environments, \textbf{Privacy Leakage} becomes a concern. \textbf{Topology Matters} \citep{LiuEtAl2025Topology} highlights that the network structure of multi-agent systems intrinsically governs memory leakage, finding that fully connected graphs maximize vulnerability. Therefore, without rigorous Access Control Lists (ACLs), as implemented in \textbf{Collaborative Memory} \citep{RezazadehEtAl2025CollaborativeMemory}, agents may inadvertently retrieve cross-session memories, leaking sensitive data to unauthorized parties.

\begin{figure}[t]
\centering
\includegraphics[width=\linewidth]{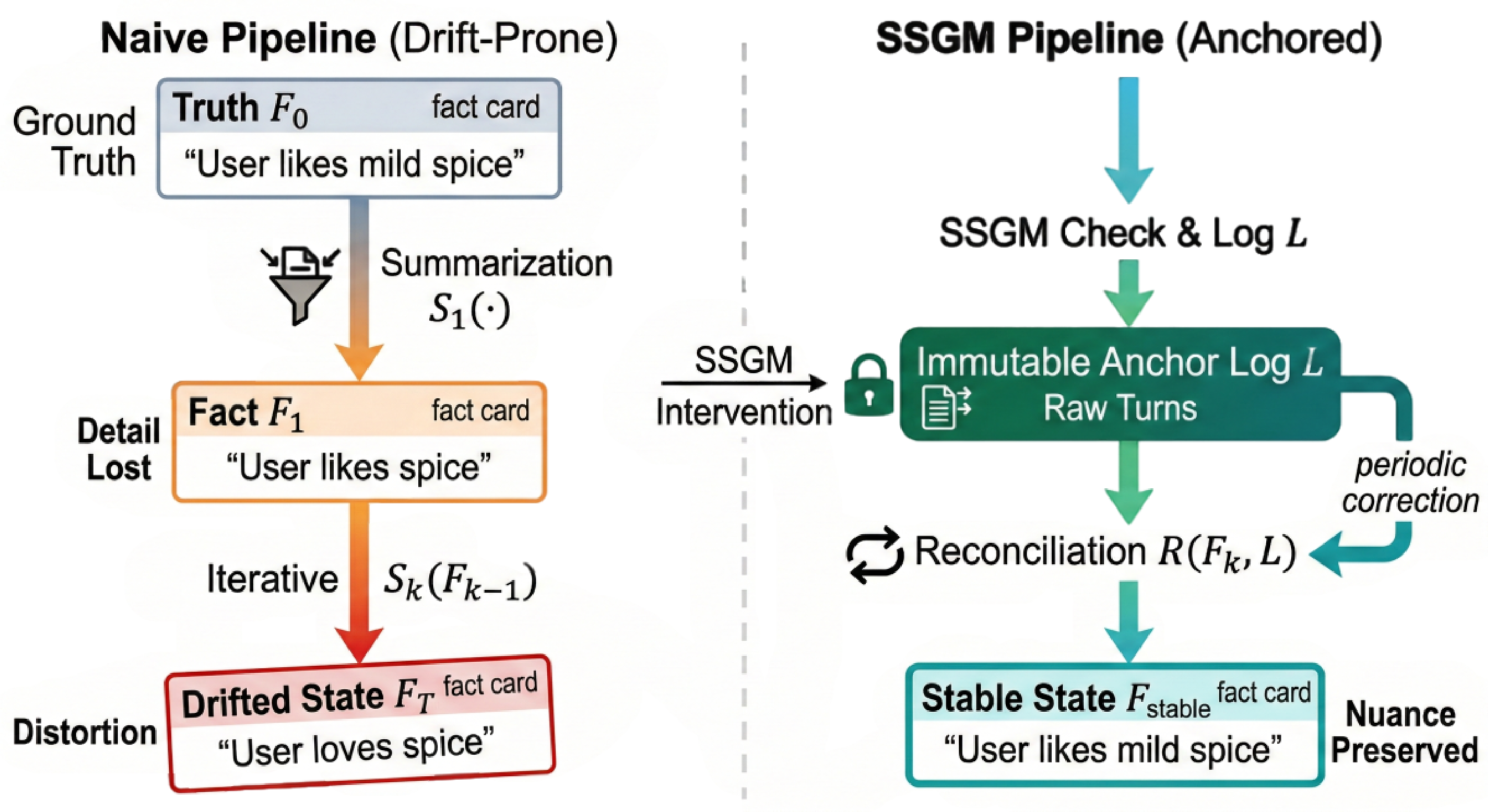}
\caption{\textbf{Logical Mechanism of Memory Drift vs. Anchored Stability.} The left path illustrates how naive iterative summarization causes semantic drift (loss of nuance). The right path demonstrates the SSGM approach, where periodic reconciliation ($\mathcal{R}$) with an immutable anchor log ($\mathcal{K}_{\text{ledger}}$) bounds the distortion relative to the ground truth ($\mathcal{K}_{\text{true}}$).}
\label{fig:drift_logic}
\end{figure}

\section{Stability and Safety Governed Memory (SSGM): Design Principles and Research Agenda}

Current memory systems predominantly prioritize adaptability---the ability of an agent to rapidly learn and incorporate new context---over stability and safety. In unrestricted architectures, the agent interacts directly with the storage medium, acting as both the sole generator and validator of its evolving knowledge base. As established in Section 4, this unconstrained autonomy is the primary catalyst for semantic drift, catastrophic forgetting, and susceptibility to adversarial memory poisoning \citep{GreshakeEtAl2023Injection}.

\begin{figure*}[t]
\centering
\includegraphics[width=\linewidth]{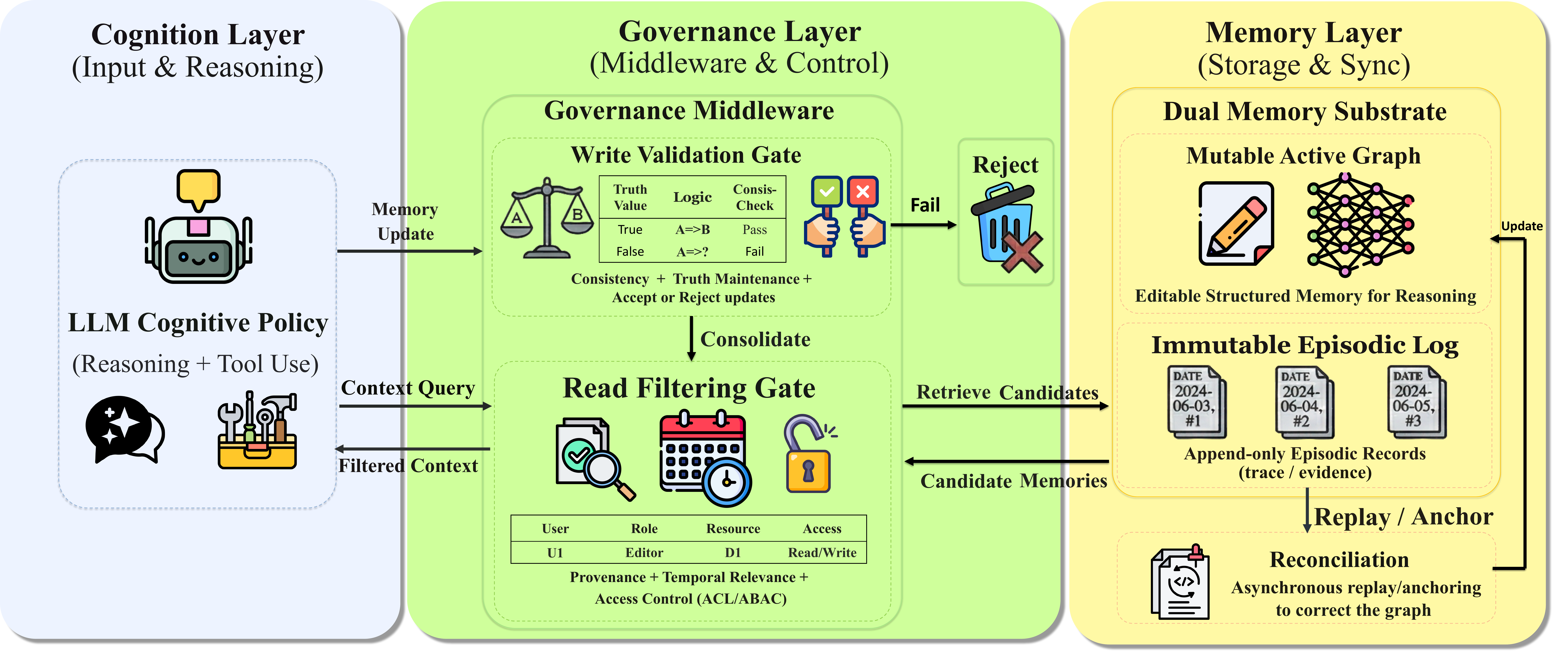}
\caption{\textbf{The conceptual architecture of the SSGM framework, featuring a decoupled left-to-right processing pipeline.} The Governance Middleware intercepts memory interactions between the LLM Agent and the Memory Substrate. The upper pathway governs memory consolidation (Write Validation) to ensure logical consistency and mitigate drift, while the lower pathway manages memory retrieval (Read Filtering) by enforcing access scopes and temporal relevance. The right-side dual-memory substrate further supports reversible reconciliation by pairing a mutable active graph with an immutable episodic log.}
\label{fig:ssgm}
\end{figure*}

To bridge this critical gap, we formulate the \textbf{Stability- and Safety-Governed Memory (SSGM)} framework. Rather than a specific software implementation, SSGM provides a rigorous theoretical architecture and a set of design principles. At its core, SSGM proposes that robust memory systems must structurally decouple the generative cognitive policy of the agent from the underlying memory substrate through an active, intercepting \textit{Governance Middleware}. Figure~\ref{fig:ssgm} operationalizes the mechanism abstracted in Figure~\ref{fig:drift_logic} as a full governed architecture, where Eq.~(5), Eq.~(6), and Eq.~(7) correspond respectively to constrained retrieval, gated writing, and asynchronous reconciliation.

\subsection{Design Principles of Governed Memory}
As illustrated in Figure~\ref{fig:ssgm}, SSGM replaces direct memory access with a set of governed gates. We articulate four foundational design principles necessary for engineering reliable, long-horizon agents:

\paragraph{Principle 1: Pre-Consolidation Validation}
In biological systems, the hippocampus does not instantly commit every perception to the neocortex; information undergoes significant filtering and reality monitoring \citep{TrespEtAl2023TensorBrain}. Similarly, in SSGM, memory updates should never be committed passively. When the agent generates a memory delta ($\Delta M$), it must pass through a \textit{Write Validation Gate} (depicted as the upper consolidation pathway in Figure~\ref{fig:ssgm}). This mechanism acts as a Truth Maintenance System (TMS) that retrieves established core facts ($M_{\text{core}}$) to execute a strict logical contradiction check: if the condition $\Delta M \wedge M_{\text{core}} \models \bot$ evaluates to true, the update is rejected. By bounding updates with formal Natural Language Inference (NLI) checks, the system actively prevents hallucination cascades from permanently corrupting the semantic graph \citep{XuEtAl2025AMem}.

\paragraph{Principle 2: Temporal and Provenance Grounding}
Not all stored information retains its validity indefinitely. Unrestricted retrieval often pulls stale or maliciously injected instructions, leading to temporal obsolescence \citep{RasmussenEtAl2025Zep}. SSGM advocates for a \textit{Read Filtering Gate} (illustrated in the lower retrieval pathway in Figure~\ref{fig:ssgm}) that evaluates candidate contexts based on two axes. First, it requires cryptographic provenance ($\sigma(\mu)$) to ensure the memory unit was generated by a trusted source rather than an adversarial prompt. Second, it applies a cognitive decay function, such as the Weibull distribution $w(\Delta \tau) = \exp(-(\Delta \tau/\eta)^\kappa)$ proposed by \citet{HuangEtAl2025LiCoMemory}. Memories falling below a dynamic relevance threshold are pruned before reaching the agent's context window, ensuring the agent reasons exclusively over fresh, high-confidence data.

\paragraph{Principle 3: Access-Scoped Retrieval}
The shift towards multi-agent \citep{WangChen2025MIRIX} and multi-tenant systems exposes critical privacy vulnerabilities. As demonstrated by \citet{LiuEtAl2025Topology}, fully connected memory networks maximize the risk of topology-induced knowledge leakage. SSGM dictates that retrieval mechanisms must not rely solely on semantic similarity. Instead, the \textit{Read Filtering Gate} must inject identity-based constraints (e.g., Attribute-Based Access Control, ABAC) directly into the query execution layer \citep{RezazadehEtAl2025CollaborativeMemory}. This ensures strict isolation of memory sub-graphs, preventing the cross-contamination of agent personas or sensitive user data.

\paragraph{Principle 4: Reversible Reconciliation}
To effectively bound long-term semantic drift, the underlying storage substrate must be dual-track. Drawing inspiration from human episodic and semantic memory divisions \citep{Alqithami2025Forgetful}, SSGM pairs a rapidly updatable \textit{Mutable Active Graph} (for fast, semantic reasoning) with an append-only \textit{Immutable Episodic Log} (acting as the operational source of truth). This dual structure enables asynchronous reconciliation, allowing the system to periodically ``replay'' and correct drifted concepts against the raw interaction trace, effectively offering a rollback mechanism if severe agent behavioral degradation occurs \citep{RathEtAl2026AgentDrift}.

\subsection{Formalizing the SSGM Framework}
To provide a rigorous theoretical account of SSGM, we formulate the agent's memory evolution as a controlled state transition. Let $M_{t-1}$ denote the mutable memory state at time $t-1$, let $M_{\text{core}}$ denote the set of protected core facts used for contradiction checking, let $\mathcal{K}_{\text{ledger}}$ denote the append-only immutable ledger of raw observations, and let $\mathcal{K}_{\text{true}}$ denote the idealized semantic target that the system seeks to approximate.The read--write lifecycle in SSGM is then governed by constrained retrieval, gated writing, and periodic reconciliation.

\textbf{The Read Phase (Constrained Retrieval):} Unlike naive Retrieval-Augmented Generation (RAG), which relies only on semantic Top-$K$ retrieval, the retrieved context $C_t$ given a query $q_t$ in SSGM is defined by jointly enforcing semantic relevance, access control, and temporal freshness:
\begin{equation}
\begin{split}
C_t = \left\{ \mu \in \text{Top-}K(q_t, M_{t-1}) \mid \mathrm{ACL}(\mu, u_{\mathrm{id}}) \right. \\
\left. \wedge \left( w(\Delta \tau_\mu) \ge \theta_{\text{fresh}} \right) \right\}
\end{split}
\end{equation}
where $\mu$ denotes a candidate memory unit, $\text{Top-}K(q_t,M_{t-1})$ denotes the set of the top-$K$ semantically retrieved items from the current memory state with respect to query $q_t$, $u_{\mathrm{id}}$ denotes the identity of the requesting user or agent, $\mathrm{ACL}(\mu,u_{\mathrm{id}})$ denotes an access-control predicate that returns true only when $u_{\mathrm{id}}$ is permitted to read memory unit $\mu$, $\Delta \tau_\mu$ denotes the elapsed time since the last valid use of $\mu$, and $\theta_{\text{fresh}}$ denotes the global freshness threshold. Intuitively, Eq.~(5) implements a two-stage read path: semantic retrieval first proposes candidates, and governance then filters them to prevent both privacy leakage and stale-memory activation.

\textbf{The Write Phase (Gated Transition):} In an unconstrained system, the next memory state would be produced by directly unioning the agent-generated update with the existing memory, i.e., $M_t = M_{t-1} \cup \mathrm{Agent}(C_t)$. In SSGM, this transition is intercepted by a write-governance operator:
\begin{equation}
M_t = M_{t-1} \cup \mathcal{G}_{\text{write}}\left(\mathrm{Agent}(C_t), M_{\text{core}}\right)
\end{equation}
where $\mathrm{Agent}(C_t)$ denotes the candidate memory delta proposed by the agent after reasoning over $C_t$, and $\mathcal{G}_{\text{write}}(\Delta M,M_{\text{core}})=\Delta M$ if $\Delta M \wedge M_{\text{core}} \not\models \bot$, and $\emptyset$ otherwise. Here, $\Delta M$ denotes the candidate update set and $\models \bot$ denotes the logical entailment of contradiction. Intuitively, Eq.~(6) turns memory writing into a guarded transition: the system only admits updates that do not conflict with protected core facts, thereby preventing hallucinated or inconsistent statements from being consolidated into long-term memory.

\textbf{Reconciliation (Drift Bounding):} To prevent the cumulative drift of admitted updates from diverging as $T \to \infty$, the asynchronous reconciliation operator $\mathcal{R}$ periodically re-aligns the mutable memory against the immutable ledger:
\begin{equation}
M_{\text{clean}} \leftarrow \arg\min_{M} \mathbb{E} \Big[ \delta \Big( \mathcal{R}(M, \mathcal{K}_{\text{ledger}}), \mathcal{K}_{\text{true}} \Big) \Big]
\end{equation}
where $M$ ranges over candidate cleaned memory states, $\mathcal{R}(M,\mathcal{K}_{\text{ledger}})$ denotes the reconciled memory obtained by replaying or correcting $M$ using the immutable ledger, $\delta(\cdot,\cdot)$ is the semantic-drift measure defined in Eq.~(4), and the expectation is taken over the stochasticity of task trajectories, noisy summaries, or reconciliation decisions. Intuitively, Eq.~(7) states that the system repeatedly searches for a cleaned memory state $M_{\text{clean}}$ whose reconciled form is, in expectation, as close as possible to the desired semantic target. This objective makes the distinction among the three reference objects explicit: $M_{\text{core}}$ protects critical facts during writing, $\mathcal{K}_{\text{ledger}}$ is the operational raw trace used for correction, and $\mathcal{K}_{\text{true}}$ is the ideal target used for evaluation.

\vspace{5pt}
\noindent\textbf{Theorem 1 (Bounded Semantic Drift).} \textit{Assume that each valid summarization or consolidation step introduces at most $\epsilon_{\text{step}}$ semantic error before reconciliation, and assume that the reconciliation operator $\mathcal{R}$ restores the mutable memory to a state whose residual error is bounded by a constant independent of the total horizon. In a naive system, the expected drift at time $T$ scales as $O(T \cdot \epsilon_{\text{step}})$. Under the SSGM framework, if reconciliation is executed every $N$ steps, the expected semantic drift is upper-bounded by $O(N \cdot \epsilon_{\text{step}})$, ensuring stability even when $T \gg N$.}

\noindent\textit{Proof Sketch.} In an unconstrained update process without Eq.~(7), per-step errors accumulate additively, yielding $\mathbb{E}[\delta(M_T)] \le \sum_{t=1}^{T} \epsilon_{\text{step}} = T \cdot \epsilon_{\text{step}}$. Under SSGM, reconciliation is invoked every $N$ steps, so error can accumulate for at most one reconciliation window before being corrected. For any $T = mN + r$ with $0 \le r < N$, the unreconciled portion contributes at most $r \cdot \epsilon_{\text{step}} < N \cdot \epsilon_{\text{step}}$, up to the bounded residual reconciliation error. Hence the dominant growth term is bounded by the window size $N$, not by the full horizon $T$.
\vspace{5pt}

\subsection{Testable Research Hypotheses and Evaluation Protocols}
The formalization of SSGM establishes a new paradigm for evaluating agent memory. Moving beyond simplistic metrics like Recall@K \citep{MemoryBench2026}, we propose three testable research hypotheses that should guide future empirical evaluations in the NLP community:

\textbf{H1: Governance Gates Statistically Bound Drift Magnitude.}
Based on Theorem 1, we hypothesize that agents equipped with $\mathcal{G}_{\text{write}}$ and $\mathcal{R}$ will exhibit an asymptotic upper limit on drift magnitude ($\delta(M_T, \mathcal{K}_{\text{true}})$) over infinite-horizon tasks (e.g., $T > 100$ turns), whereas baseline agents will demonstrate approximately linear drift accumulation. \textit{Evaluation Protocol:} This can be robustly tested using established benchmarks like \textit{LongMemEval} \citep{wu2025longmemeval}, utilizing LLM-as-a-Judge and metrics like BERTScore to quantify the fidelity of the final memory graph against the ground-truth text over extended timesteps.

\textbf{H2: Access-Scoped Retrieval Lowers Adversarial Leakage Risk.}
We hypothesize that enforcing the constrained retrieval rule in Eq.~(5) will dramatically lower the successful injection rate of cross-tenant adversarial prompts without degrading the primary task success rate. \textit{Evaluation Protocol:} Researchers should simulate multi-user role-playing scenarios and measure the ``Leakage Plateau'' \citep{LiuEtAl2025Topology} when adversarial data is deliberately injected into neighboring graph nodes.

\textbf{H3: The Latency vs. Coherence Trade-off.}
Integrating strict logical contradiction checks into the critical path of Eq.~(6) will yield a measurable increase in memory write latency. \textit{Evaluation Protocol:} Future studies must quantify this trade-off, potentially demonstrating that \textit{asynchronous} governance protocols (e.g., executing $\mathcal{R}$ during idle periods) can achieve high coherence (low $\delta$) without compromising the immediate conversational fluidity \citep{ZhangEtAl2025aGMemory}.

\section{Conclusion}
The transition from static context windows to evolving, self-refining memory represents a pivotal leap in agentic AI. However, as demonstrated in this survey, this autonomy comes with the peril of \textbf{memory corruption}---ranging from subtle semantic drift to catastrophic poisoning. We argue that the prevailing focus on ``retrieval accuracy'' is insufficient; the next generation of memory systems must prioritize \textbf{memory integrity} and \textbf{safety}.

The \textbf{SSGM framework} proposed in this paper serves as a conceptual foundation for rigorous memory governance. By decoupling the cognitive policy from the memory substrate via validation and filtering gates, we can build agents that are adaptable yet robust. Beyond proposing SSGM, this work also provides a taxonomy of evolving memory systems, a structured analysis of failure modes, and a framing of the core trade-offs that will shape future governed memory designs. Looking forward, the community must focus on standardized safety benchmarks (like \textit{MemoryBench}) that stress-test memory stability under adversarial drift, machine unlearning protocols to surgically remove toxic memories, and evaluating the hypotheses defined in our research agenda. Ultimately, solving the memory-governance problem is a prerequisite for deploying lifelong learning agents in high-stakes, real-world environments.

\section*{Limitations}
While SSGM provides a blueprint for reliable memory, it introduces three fundamental trade-offs that warrant future investigation:

\textbf{1. The Latency-Safety Trade-off:} The proposed governance layer introduces a ``System 2'' verification step into the memory loop. Validating consistency and provenance for every update incurs significant latency, potentially rendering the agent unresponsive in real-time scenarios. Future work must explore \textbf{asynchronous governance}, where memory is optimistically updated but periodically ``sanitized'' in the background \citep{ZhangEtAl2025aGMemory}.

\textbf{2. The Stability-Plasticity Conflict:} Strict consistency filtering may lead to \textbf{knowledge ossification}. If the governance layer aggressively rejects information that conflicts with established memory, the agent may fail to adapt to legitimate environmental changes (e.g., a user changing their address). Designing ``conflict resolution protocols'' that can distinguish between \textit{drift} and \textit{update} remains an open algorithmic challenge.

\textbf{3. Scalability of Graph Structures:} While graph-based memories (e.g., Zep, MAGMA) offer superior reasoning capabilities, maintaining a consistent graph at scale is non-trivial. As interaction history grows, the complexity of graph traversal and entity resolution can degrade retrieval performance, necessitating more efficient graph pruning and compression algorithms \citep{JiangEtAl2026MAGMA}.

\bibliographystyle{acl_natbib}
\bibliography{ssgm}

\end{document}